\documentclass[11pt]{article}

\usepackage[preprint]{acl}

\usepackage{times}
\usepackage{latexsym}

\usepackage[T1]{fontenc}

\usepackage[utf8]{inputenc}

\usepackage{microtype}

\usepackage{inconsolata}

\usepackage{graphicx}
\usepackage[ruled,linesnumbered,vlined]{algorithm2e}
\usepackage{algpseudocode}
\usepackage{hyperref}
\usepackage{url}
\usepackage{amsmath}
\usepackage{amsfonts}
\usepackage{subcaption}
\usepackage{bm}
\usepackage{booktabs}
\usepackage{caption}
\usepackage{multirow}
\usepackage{array}
\usepackage{enumitem}
\usepackage{cleveref}
\usepackage{stfloats}
\usepackage{listings}
\usepackage{xcolor}
\usepackage{tcolorbox}
\usepackage{colortbl}
\usepackage{textcomp}
\usepackage{tabularx}
\usepackage{dashrule}
\usepackage{relsize}

\definecolor{lavender}{RGB}{230,230,250}
\definecolor{brickred}{RGB}{150,0,24}

\colorlet{punct}{red!60!black}
\definecolor{background}{HTML}{F5F5F5}
\definecolor{delim}{RGB}{20,105,176}
\colorlet{numb}{magenta!60!black}

\lstdefinelanguage{json}{
    basicstyle=\normalfont\ttfamily\small,
    numbers=left,
    numberstyle=\scriptsize,
    stepnumber=1,
    numbersep=8pt,
    showstringspaces=false,
    breaklines=true,
    frame=lines,
    backgroundcolor=\color{background},
    literate=
     *{0}{{{\color{numb}0}}}{1}
      {1}{{{\color{numb}1}}}{1}
      {2}{{{\color{numb}2}}}{1}
      {3}{{{\color{numb}3}}}{1}
      {4}{{{\color{numb}4}}}{1}
      {5}{{{\color{numb}5}}}{1}
      {6}{{{\color{numb}6}}}{1}
      {7}{{{\color{numb}7}}}{1}
      {8}{{{\color{numb}8}}}{1}
      {9}{{{\color{numb}9}}}{1}
      {:}{{{\color{punct}{:}}}}{1}
      {,}{{{\color{punct}{,}}}}{1}
      {\{}{{{\color{delim}{\{}}}}{1}
      {\}}{{{\color{delim}{\}}}}}{1}
      {[}{{{\color{delim}{[}}}}{1}
      {]}{{{\color{delim}{]}}}}{1},
}

\title{\textsc{R$^{2}$Adapter}: A \underline{R}outing and \underline{R}ewriting \underline{Adapter} for Efficient Hybrid RAG}

\author{
    Yucan Guo\textsuperscript{1,2,3},
    Miao Su\textsuperscript{1,2,3},
    Saiping Guan\textsuperscript{1,2,3}\thanks{Corresponding authors.},
    Long Bai\textsuperscript{1,2,3},
    Zhongni Hou\textsuperscript{4}, \\
    \textbf{Zixuan Li\textsuperscript{1,2,3},
    Xiaolong Jin\textsuperscript{1,2,3}$^{\dagger}$,
    Jiafeng Guo\textsuperscript{1,2,3},
    Xueqi Cheng\textsuperscript{1,2,3}}
     \\
    \textsuperscript{1}State Key Laboratory of AI Safety \\
    \textsuperscript{2}Institute of Computing Technology, Chinese Academy of Sciences \\
    \textsuperscript{3}University of Chinese Academy of Sciences\\
    \textsuperscript{4}Meituan\\
    \small
    \texttt{\{guoyucan23z, guansaiping, jinxiaolong\}@ict.ac.cn}\\
}

\newcommand{\revise}[1]{\textcolor{black}{#1}}

\newcommand{\method}{\textsc{R$^{2}$Adapter}}

\begin{document}
\maketitle
\begin{abstract}
Retrieval-Augmented Generation (RAG) has become a prevailing paradigm for enhancing Large Language Models (LLMs) with non-parametric knowledge. 
Vanilla RAG efficiently handles simple queries but struggles with relational or multi-hop reasoning. Graph-based RAG alleviates this issue but incurs higher inference complexity and latency.
In practice, user queries can differ significantly in their complexity, rendering a fixed RAG strategy suboptimal.
However, existing hybrid text-graph RAG methods typically rely on heuristic and LLM-based routing, resulting in unnecessary overhead and strong dependence on the underlying LLM.
To address these challenges, we propose \method{}, a lightweight plug-in \textbf{R}outing and \textbf{R}ewriting \textbf{Adapter} designed to allocate queries between vanilla and graph-based RAG dynamically. By routing only the queries that genuinely benefit from graph-based reasoning, \method{} reduces unnecessary graph retrieval overhead. Additionally, uncertain graph-routed queries are rewritten to better expose their multi-hop reasoning requirements, improving retrieval quality without additional supervision.
Extensive experiments on three multi-hop QA benchmarks demonstrate that \method{} reduces graph-based RAG usage by up to 59\% while maintaining comparable answer accuracy. This adapter~\footnote{The code is publicly available at \url{https://github.com/YucanGuo/R2Adapter}.} is model-agnostic and can be seamlessly integrated into diverse vanilla and graph-based RAG pipelines, providing an efficient and adaptive solution for hybrid RAG systems.
\end{abstract}

\section{Introduction}
\label{sec:Introduction}
Large Language Models (LLMs) have achieved strong performance across a wide range of natural language processing tasks~\citep{Zhao2023survey, Team2024gemini, Guo2025deepseek, Qwen2025qwen25}. However, their reliance on parametric knowledge limits factual correctness when handling knowledge-intensive queries or information beyond their training distribution~\citep{Augenstein2024factuality, Huang2025survey}. Retrieval-Augmented Generation (RAG) mitigates this limitation by incorporating non-parametric external knowledge into LLMs, enabling models to generate their responses based on the retrieved evidence and improving both accuracy and interpretability~\citep{Lewis2020retrieval, Gao2023retrieval, Zhao2024retrieval}.

\begin{figure}[t]
    \centering
    \includegraphics[width=.9\linewidth]{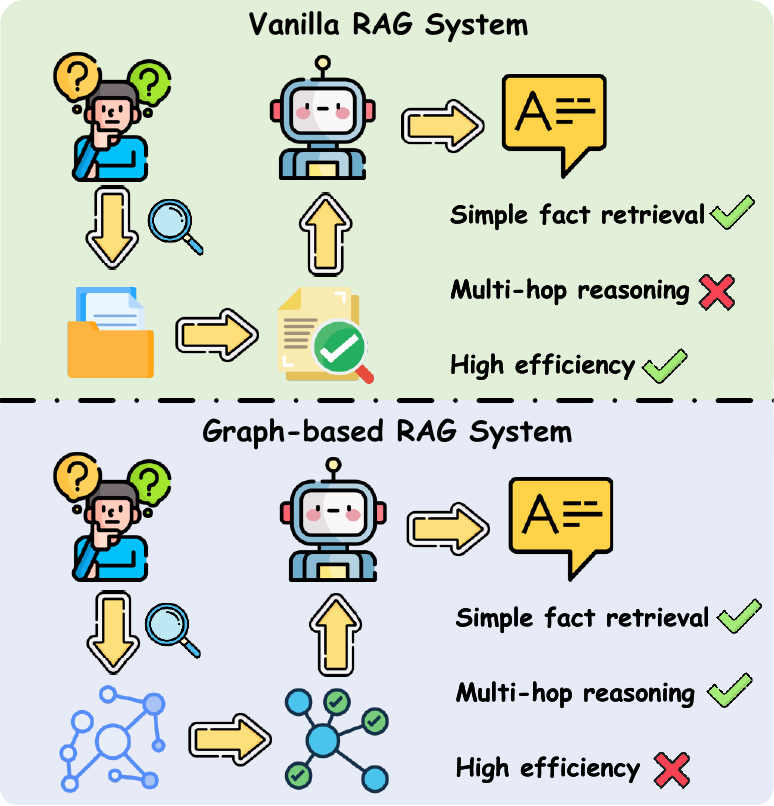}
    \caption{Vanilla RAG vs. Graph-based RAG.}
    \label{fig:vanilla_vs_graphrag}
\end{figure}

As illustrated in~\Cref{fig:vanilla_vs_graphrag}, existing RAG methods largely follow two paradigms, i.e., vanilla RAG and graph-based RAG. Vanilla RAG retrieves unstructured passages and integrates them into the generation process, providing an efficient solution for queries that can be resolved through single-hop retrieval or surface-level evidence matching~\citep{Guu2020retrieval, Karpukhin2020dense, Wang2023self}. However, it struggles with queries requiring multi-hop inference or compositional evidence aggregation across multiple entities and facts.
In contrast, graph-based RAG organizes unstructured passages into structured knowledge graphs~\citep{Edge2024local, Gutierrez2024hipporag, Gutierrez2025from}, explicitly modeling entities and relations to support complex and multi-hop reasoning~\citep{Peng2025graph, Zhou2025depth, Zhang2025survey}. Despite these advantages, graph-based RAG incurs high inference latency and computational overhead due to graph construction and retrieval, which limits its scalability and practical deployment.

In real-world settings, user queries vary widely in their reasoning requirements, ranging from simple fact retrieval to complex multi-hop relational inference. To accommodate this diversity, a natural solution is to combine vanilla and graph-based RAG within a hybrid framework, dynamically selecting the most suitable RAG paradigm for each query. Existing hybrid text-graph RAG approaches typically rely on heuristic rules, such as the number of entities in a query~\citep{Lee2025hybgrag}, or LLM-based query classifiers~\citep{Han2025rag}. However, these approaches face notable limitations. Heuristic strategies are often brittle and fail to generalize across queries, while LLM-based routing incurs substantial computational cost and inference latency. In practice, invoking an LLM solely for query classification can be as expensive as, or even more costly than, directly applying graph-based RAG. Consequently, effective hybrid RAG systems require routing mechanisms that are not only adaptive but also lightweight, fast, and cost-efficient.

To address these challenges, we propose \textbf{\method{}}, a plug-in \textbf{R}outing and \textbf{R}ewriting \textbf{Adapter} designed to enable efficient hybrid RAG. The core purpose of \method{} is to utilize the complementary strengths of vanilla and graph-based RAG while avoiding their respective inefficiencies. Rather than committing to a fixed retrieval strategy or relying on costly LLM-based routing, \method{} performs as a plug-in module that can be seamlessly integrated into existing RAG systems.
\method{} consists of two components: (1) a lightweight router is trained to rapidly estimate whether a given query is likely to benefit from graph-based reasoning. The router operates as a low-cost classifier for query complexity assessment, enabling fast strategy selection without invoking LLMs; and (2) a query rewriter that selectively refines queries with low routing confidence for graph-based RAG to better unveil its latent relational structure, ensuring effective graph reasoning even when queries are ambiguous.
\method{} is model-agnostic, requiring no modification to the underlying vanilla or graph-based RAG systems. As a result, it provides a practical and general solution that improves the efficiency of graph-based RAG while preserving its reasoning benefits, making it well-suited for scalable and real-world hybrid RAG deployments.

Our contributions are summarized as follows:
\begin{itemize}[leftmargin=*]
    \item We propose \method{}, a plug-in adapter for hybrid RAG systems that dynamically routes queries between vanilla and graph-based RAG, and selectively rewrites queries to enhance graph-based RAG further.
    \item To enable fast and low-cost routing, we construct a training corpus that characterizes whether queries can be successfully handled by vanilla RAG or require graph-based reasoning, and train a lightweight router to predict retrieval strategy without relying on LLMs.
    \item Extensive experiments on multi-hop QA benchmarks show that \method{} significantly reduces graph-based RAG usage while preserving answer accuracy across representative graph-based RAG systems, demonstrating its effectiveness and generality.
\end{itemize}

\begin{figure*}
    \centering
    \includegraphics[width=\linewidth]{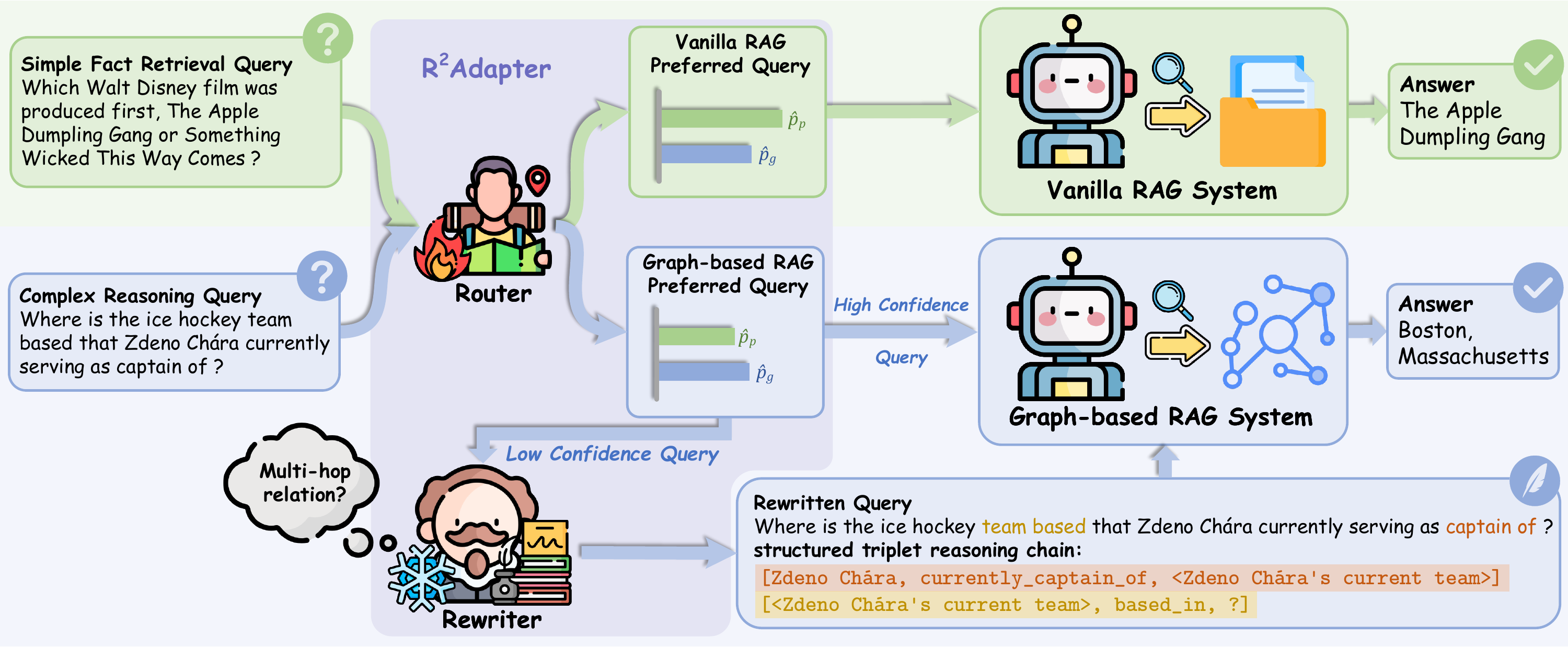}
    \caption{The overview of how \method{} is integrated with vanilla and graph-based RAG systems.}
    \label{fig:method}
\end{figure*}

\section{Related Work}
\label{sec:Related Work}
\textbf{Graph-based RAG.}
RAG enhances LLMs with external knowledge, where early approaches predominantly rely on unstructured passage retrieval~\citep{Lewis2020retrieval, Karpukhin2020dense, Jiang2023active, Asai2024self} and often struggle with relational and multi-hop queries. To address this limitation, recent work augments RAG with structured graphs that enable explicit reasoning over entities and relations~\citep{Peng2025graph, Han2024retrieval}.
RAPTOR~\citep{Sarthi2024raptor} organizes documents into a hierarchical summary tree for retrieval. Moving to explicit graph structures with entities and relations, GraphRAG~\citep{Edge2024local} builds a knowledge graph with community summaries and performs retrieval via graph summarization, while LightRAG~\citep{Guo2025lightrag} improves efficiency through simplified indexing and fast dual-level retrieval. 
HippoRAG~\citep{Gutierrez2024hipporag} builds a memory-inspired knowledge graph and retrieves via personalized PageRank, and HippoRAG 2~\citep{Gutierrez2025from} extends HippoRAG with deeper passage integration.
Despite their reasoning benefits, graph-based RAG methods incur a higher computational cost and are unnecessary for many simple queries that can be effectively resolved by vanilla RAG.

\smallskip
\noindent \textbf{Hybrid RAG.}
Hybrid RAG is motivated by the observation that user queries exhibit varying levels of complexity, and that no single retrieval strategy is uniformly optimal across all queries. Early hybrid frameworks mainly focus on text retrieval~\citep{Mallen2023when, Jeong2024adaptive}. For instance, Adaptive-RAG~\citep{Jeong2024adaptive} employs a classifier to dynamically route queries among single-step text retrieval, multi-step text retrieval, or non-retrieval pathways.
With the advent of graph-based RAG, recent work has explored hybrid frameworks that combine unstructured text retrieval with structured graph-based retrieval. HyPA-RAG~\citep{Kalra2025hypa} integrates text and graph evidence directly, while HybGRAG~\citep{Lee2025hybgrag} selectively activates textual or graph retrieval depending on whether relevant entities can be extracted from the query by an LLM. Similarly, \citet{Han2025rag} adopts an LLM-based classifier to route queries to text or graph retrieval, depending on whether the query is fact-based or reasoning-based.
However, existing hybrid text-graph RAG methods either always invoke graph retrieval or rely on LLM-based classifiers for routing, which incur substantial computational overhead and are tightly coupled to the capability of the underlying LLM.

\section{\method{}}
\label{sec:Method}
In this section, we present \method{}, a plug-in adapter consisting of a router and a rewriter. 
In~\S\ref{sec:Overall Framework}, we provide an overview of the framework.
Then, we introduce the router and the rewriter in~\S\ref{sec:Router} and~\S\ref{sec:Rewriter}, respectively.

\newcommand{\algcommentstyle}[1]{\ttfamily\textcolor{blue}{#1}}
\SetCommentSty{algcommentstyle}

\begin{algorithm}[t]
\caption{\method{}}
\small
\label{alg:method_overview}
\KwIn{
User query $q$, Router $\mathcal{R}$, Rewriter $\mathcal{W}$, \\
Vanilla RAG system $\texttt{RAG}_{p}$, and Graph-based RAG system $\texttt{RAG}_{g}$.
}
\KwOut{
Generated answer $a$.
}

$r \gets \mathcal{R}(q)$\tcp*{Routing}
\eIf{$r = \textsc{Passage}$}{
    \tcc{Passage-based vanilla RAG}
    $a \gets \texttt{RAG}_{p}(q)$\;
}{
    \tcc{Graph-based RAG}
    \If{$\mathcal{R}$ is uncertain on $q$}{
        $q' \gets \mathcal{W}(q)$\tcp*{Rewriting}
    }
    \Else{
        $q' \gets q$\;
    }
    $a \gets \texttt{RAG}_{g}(q')$\;
}
\Return{$a$}\;
\end{algorithm}

\subsection{Overall Framework}
\label{sec:Overall Framework}
\Cref{fig:method}{} illustrates the overall architecture of \method{}. Given an input user query $q$, \method{} operates as a lightweight pre-retrieval adapter that determines how the query should be processed by the downstream RAG systems. The adapter consists of two sequential components: a \textit{router} that predicts the preference of vanilla and graph-based RAG for the query, and a \textit{rewriter} that selectively refines queries routed to graph-based RAG under uncertainty.
Specifically, the router takes the raw query as input and produces a routing decision indicating whether the query is likely to benefit from vanilla or graph-based retrieval. 
Queries predicted to be solvable by vanilla RAG are directly forwarded to a standard passage-based RAG pipeline, avoiding unnecessary graph traversal.
For queries routed to graph-based RAG, \method{} further considers the confidence of the routing decision. When the router exhibits low confidence, the query is passed to a query rewriter that reformulates the original input to show its latent relational structure better. The rewritten query is then used for graph-based retrieval and subsequent answer generation. The algorithmic description of \method{} is shown in~\Cref{alg:method_overview}.

\begin{figure}[t]
    \centering
    \includegraphics[width=\linewidth]{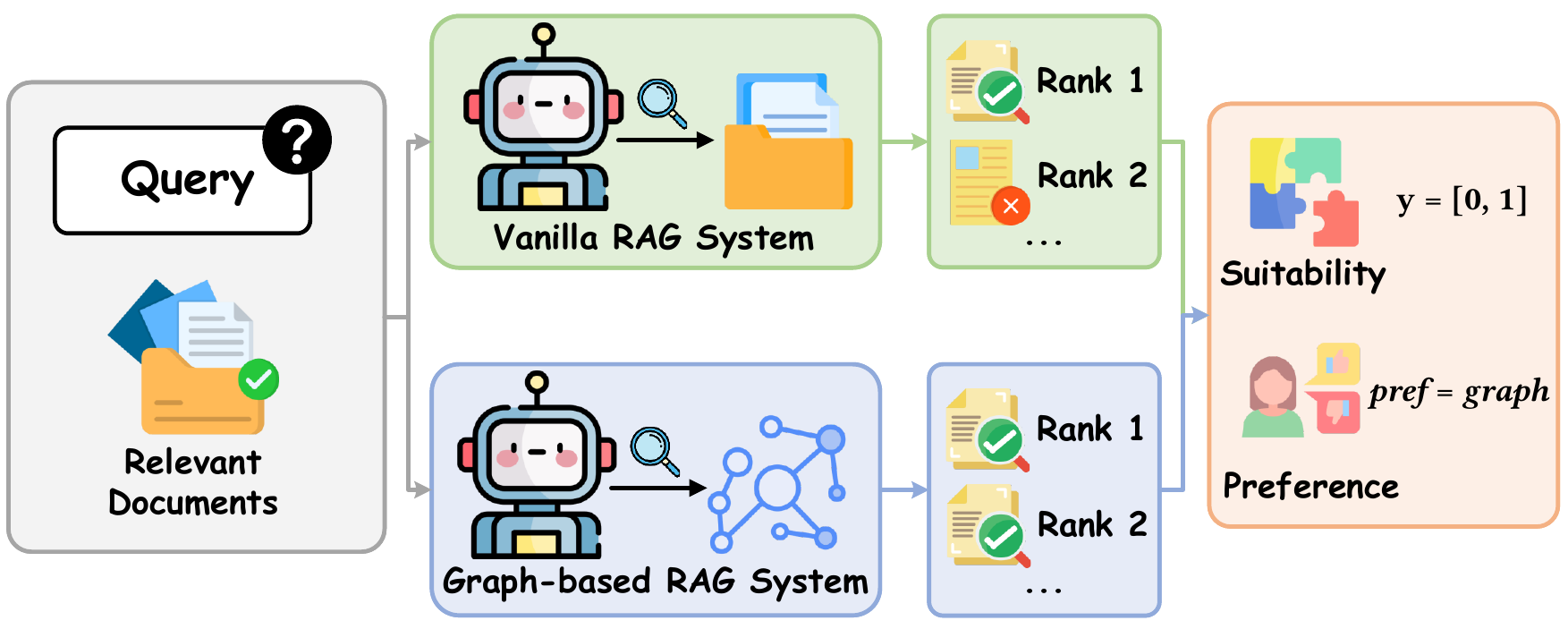}
    \caption{Automatic annotation pipeline for router training corpus.}
    \label{fig:annotation}
\end{figure}

\subsection{The Router}
\label{sec:Router}
We present a lightweight router that predicts the preference of vanilla and graph-based RAG for each query. The following sections describe the construction of its training corpus, along with the details of the training and routing process.
\subsubsection{Training Corpus Construction}
\label{sec:Training Corpus Construction}
We design an automatic annotation pipeline, as illustrated in~\Cref{fig:annotation}, to construct the training corpus for the router.
The corpus is built on the ``Hard'' subset of HotpotQA~\citep{Yang2018hotpotqa}.
For each query, we evaluate two RAG paradigms: (1) a vanilla RAG system employing passage-based retrieval~\citep{Lewis2020retrieval} and using NV-Embed-v2~\citep{Lee2025nvembed} as the retriever; and (2) a graph-based RAG system using HippoRAG 2~\citep{Gutierrez2025from}. Each system retrieves a ranked list of documents, from which we compute the average rank of the relevant documents as a measure of retrieval quality.

Based on this criterion, we assign a dual-dimensional suitability label $\mathbf{y} = [y_p, y_g]$, where $y_p, y_g \in \{0,1\}$ indicate whether vanilla RAG or graph-based RAG is suitable for the query, respectively. The retrieval paradigm with the lower average document rank is marked as suitable. When both paradigms achieve identical average ranks, both are considered suitable, yielding the label $[1,1]$.
Based on suitability labels, we further define a preference label $pref \in \{passage, graph\}$ that specifies which paradigm should be selected. If only one paradigm is suitable, it is preferred by default. When both paradigms are suitable, preference is assigned to vanilla RAG for its lower computational cost and inference latency.

Table~\ref{tab:router_data_stats} reports the statistics of the constructed corpus. While the majority of queries can be effectively handled by either paradigm, a non-trivial subset exhibits a clear preference for graph-based RAG.

\subsubsection{Training and Routing Process}
\begin{table}[t]
\centering
\caption{Statistics of the training corpus. A query is considered recalled if all relevant documents are retrieved within the top 5.}
\label{tab:router_data_stats}
\small
\begin{tabular}{l r}
\toprule
\multicolumn{2}{c}{\textbf{Retrieval Coverage}} \\
\midrule
Total queries & 14{,}969 \\
Only passage-based recalled & 33 \\
Only graph-based recalled & 662 \\
Both paradigms recalled & 14{,}274 \\
\midrule
\multicolumn{2}{c}{\textbf{Routing Preference}} \\
\midrule
Preference for vanilla RAG & 87.8\% \\
Preference for graph-based RAG & 12.2\% \\
Both paradigms suitable & 81.9\% \\
\bottomrule
\end{tabular}
\end{table}
\textbf{Training Process.}
Given the constructed training corpus, we train the router as a dual-output classifier that independently estimates the suitability of vanilla and graph-based RAG for each query. The router adopts a DeBERTa-v3-base~\citep{He2023deberta} encoder as its backbone, followed by a lightweight classifier that produces two outputs, corresponding to the score of two RAG paradigms.

Formally, given a query $q$ with suitability label $\mathbf{y}=[y_p, y_g]$, the router first encodes the query into a representation $\mathbf{h}_q$, which is then mapped to two suitability scores through a linear layer and sigmoid activation:
\begin{equation}
\hat{\mathbf{p}} = [\hat{p}_p, \hat{p}_g] = \sigma(\mathbf{W}\mathbf{h}_q + \mathbf{b}),
\end{equation}
where $\hat{p}_p$ and $\hat{p}_g$ denote the predicted suitability probabilities for vanilla and graph-based RAG, respectively.

To account for class imbalance and hard examples, we adopt an independent focal Binary Cross-Entropy (BCE) loss for each output dimension. For $k \in \{p, g\}$, the loss is defined as:
\begin{equation}
\mathcal{L}_k =
\alpha_k \, (1 - p_{t,k})^{\gamma} \,
\mathrm{BCE}(\hat{p}_k, y_k),
\end{equation}
where $\mathrm{BCE}(\hat{p}_k, y_k)$ denotes the BCE loss,
$p_{t,k}$ is the predicted probability of the ground-truth class, $\alpha_k$ is a balancing factor to handle class imbalance, and $\gamma$ is the focusing parameter that down-weights easy examples.

The final training objective is computed as the average loss over the two dimensions:
\begin{equation}
\mathcal{L}_{\text{router}} = \frac{1}{2} \sum_{k \in \{p, g\}} \mathcal{L}_k.
\end{equation}

This independent formulation allows the router to model overlapping applicability between vanilla and graph-based RAG, rather than enforcing a mutually exclusive decision during training.

\smallskip
\noindent \textbf{Routing Process.}
At inference time, the router outputs suitability probabilities $\hat{\mathbf{p}}=[\hat{p}_p,\hat{p}_g]$ for vanilla and graph-based RAG, respectively. Routing decisions are made by comparing the relative suitability between the two paradigms. Specifically, a query is routed to the graph-based RAG if
\begin{equation}
\hat{p}_g - \hat{p}_p > \tau,
\end{equation}
and is otherwise routed to vanilla RAG, where $\tau$ is the routing threshold.

The routing threshold is selected on a validation set using preference annotations. In this process, graph-based RAG is treated as the positive class, reflecting the asymmetric risk between routing errors.
Routing a query that genuinely requires graph-based reasoning to vanilla RAG is particularly harmful, as the latter is unlikely to resolve such queries. In contrast, routing a passage-suitable query to graph-based RAG typically preserves answer correctness, albeit at the cost of reduced efficiency. 
As a result, recall for graph-based RAG is a critical consideration in threshold selection.
Formally, for a given threshold $\tau$, we evaluate routing performance using a balanced metric $\mathrm{Score}(\tau)$ that jointly considers F1 score and recall for the graph-based class:
\begin{equation}
\mathrm{Score}(\tau) = \frac{\mathrm{F1}(\tau) + \mathrm{Recall}(\tau)}{2}.
\end{equation}
The threshold $\tau$ that maximizes $\mathrm{Score}(\tau)$ is selected for inference, enabling a principled trade-off between overall routing accuracy and the risk of misrouting graph-required queries to vanilla RAG.

\subsection{The Rewriter}
\label{sec:Rewriter}
The query rewriter is introduced to assist graph-based RAG under routing uncertainty.
It makes the latent relational and multi-hop structure of the original query more explicit, thereby facilitating downstream reasoning.

Given an input query $q$, the rewriter aims to convert it into a reasoning-oriented representation $q_r$ that explicitly exposes its underlying relational dependencies.
Specifically, the rewriter decomposes the query into a sequence of relational statements expressed as triplets of the form
\texttt{[head\_entity, relation, tail\_entity]}.
Intermediate entities that are not explicitly mentioned in the original query will be represented as placeholders, while the final triplet will mark the target entity as ``?''.
For example, the query ``Where is the ice hockey team based that Zdeno Chára is currently serving as captain of?'' can be decomposed as:
\begin{itemize}[leftmargin=*]
    \item \texttt{[Zdeno Chára, currently\_captain\_of, <Zdeno Chára's current team>]}
    \item \texttt{[<Zdeno Chára's current team>, based\_in, ?]}
\end{itemize}

After acquiring $q_r$, it will be concatenated with the original query to form the final rewritten query:
\begin{equation}
    q' = \mathrm{concat}(q,\, q_{r}).
\end{equation}

Notably, the LLM-based rewriter $\mathcal{W}$ is activated only when the router selects graph-based RAG but exhibits low confidence in this decision:
\begin{equation}
\label{equ:rewriter threshold}
    \hat{p}_g < \epsilon,
\end{equation}
where $\epsilon$ denotes a predefined confidence threshold.
In this way, rewriting is applied conservatively, maximizing its potential benefit while minimizing unnecessary LLM calls for queries that can be effectively handled without rewriting.
The prompt template of the rewriter is provided in Appx.~\S\ref{appendix: Rewriter Prompt Template}.

\section{Experiments}
\label{sec:Experiments}
In this section, we conduct experiments to evaluate the effectiveness, efficiency, and generalization of \method{}. Our experimental study is designed to answer the following research questions:
\begin{itemize}
    \item [\textbf{RQ1}] How does \method{} perform compared to vanilla RAG and graph-based RAG?
    \item [\textbf{RQ2}] How effective is the proposed router, and how does it compare with routing strategies of other hybrid text-graph RAG systems?
    \item [\textbf{RQ3}] How does the query rewriter influence the performance of graph-based RAG under uncertain routing decisions?
\end{itemize}

\subsection{Experimental Setting}
\textbf{Evaluation Datasets.}
Following~\citet{Gutierrez2024hipporag} and~\citet{Gutierrez2025from}, we evaluate \method{} on three widely used benchmarks for multi-hop QA, namely HotpotQA~\citep{Yang2018hotpotqa}, 2WikiMultihopQA (2Wiki)~\citep{Ho2020constructing}, and MuSiQue~\citep{Trivedi2022musique}. 

\begin{table*}[htbp]
\centering
\caption{QA performance comparison. The best results are indicated in bold, while the underlined values represent the second-best results.}
\label{tab:main_results_QA}
\small
\renewcommand{\arraystretch}{1.1}
\resizebox{.85\textwidth}{!}{%
\begin{tabular}{lcccccccc}
\toprule
\multirow{2}{*}{\textbf{Method}}
 & \multicolumn{2}{c}{\textbf{HotpotQA}} 
 & \multicolumn{2}{c}{\textbf{2Wiki}} 
 & \multicolumn{2}{c}{\textbf{MuSiQue}}
 & \multicolumn{2}{c}{\textbf{Average}}\\
\cmidrule(lr){2-3} \cmidrule(lr){4-5} \cmidrule(lr){6-7} \cmidrule(lr){8-9}
 & \textbf{EM} & \textbf{F1} 
 & \textbf{EM} & \textbf{F1}
 & \textbf{EM} & \textbf{F1}
 & \textbf{EM} & \textbf{F1}\\
 \midrule
\rowcolor{gray!10}
\multicolumn{9}{c}{\textit{\textbf{Vanilla RAG}}}\\
\midrule
BM25~\citep{Robertson1994some} & 52.0 & 63.4 & 47.9 & 51.2 & 20.3 & 28.8 & 40.1 & 47.8 \\
Contriever~\citep{Izacard2022unsupervised} & 51.3 & 62.3 & 38.1 & 41.9 & 24.0 & 31.3 & 37.8 & 45.2 \\
NV-Embed-v2~\citep{Lee2025nvembed} & 61.6 & 74.7 & 57.8 & 62.5 & 34.4 & 45.0 & 51.3 & 60.7 \\
\midrule
\rowcolor{gray!10}
\multicolumn{9}{c}{\textit{\textbf{Graph-based RAG}}}\\
\midrule
LightRAG~\citep{Guo2025lightrag}  & 2.0 & 2.4 & 9.4 & 11.6 & 0.5 & 1.6 & 4.0 & 5.2 \\
RAPTOR~\citep{Sarthi2024raptor}  & 56.8 & 69.5 & 47.3 & 52.1 & 20.7 & 28.9 & 41.6 & 50.2 \\
HippoRAG~\citep{Gutierrez2024hipporag}  & 52.6 & 63.5 & 65.0 & 71.8 & 26.2 & 35.1 & 47.9 & 56.8 \\
GraphRAG~\citep{Edge2024local} & 56.6 & 70.5 & 47.1 & 53.5 & 26.3 & 37.9 & 43.3& 54.0 \\
HippoRAG 2~\citep{Gutierrez2025from} & \underline{62.5} & \underline{75.6} & \textbf{66.0} & \textbf{72.9} & \underline{37.6} & \underline{48.9} & \underline{55.4} & \underline{65.8} \\
\rowcolor{lavender}
\textbf{\method{} (GraphRAG)} & 60.2 & 74.2 & 56.8 & 62.8 & 32.1 & 43.1 & 49.7 & 60.0 \\
\rowcolor{lavender}
\textbf{\method{} (HippoRAG 2)} & \textbf{63.1} & \textbf{76.1} & \underline{65.7} & \underline{72.7} & \textbf{40.0} & \textbf{52.0} & \textbf{56.3} & \textbf{66.9} \\
\bottomrule
\end{tabular}}
\end{table*}

\smallskip
\noindent \textbf{Baselines.}
We compare \method{} against two types of approaches:  
(1) Vanilla RAG~\citep{Lewis2020retrieval} with different representative retrievers, including BM25~\citep{Robertson1994some}, Contriever~\citep{Izacard2022unsupervised}, and NV-Embed-v2~\citep{Lee2025nvembed}.  
(2) Graph-based RAG methods, including LightRAG~\citep{Guo2025lightrag}, RAPTOR~\citep{Sarthi2024raptor}, HippoRAG~\citep{Gutierrez2024hipporag}, GraphRAG~\citep{Edge2024local}, and HippoRAG 2~\citep{Gutierrez2025from}, which leverage structured graphs for retrieval. 
In addition, for router-specific analysis, we introduce two hybrid graph-text RAG baselines:
(1) NER-Router, a lightweight variant inspired by HybGRAG~\citep{Lee2025hybgrag}, which routes queries based on the number of named entities detected by spaCy~\citep{Honnibal2020spaCy}.
(2) LLM-Router~\citep{Han2025rag}, which uses an LLM to classify queries as fact-based or reasoning-based.
Both baselines are evaluated only in the router analysis and do not incorporate rewriting for a fair comparison, as these hybrid RAG methods focus solely on routing.

\smallskip
\noindent \textbf{Implementation Details.}
The router in \method{} is implemented using DeBERTa-v3-base~\citep{He2023deberta} as the backbone encoder. 
The confidence threshold $\epsilon$ of the rewriter is set to 0.6.
For all graph-based RAG methods, including those integrated with \method{}, we adopt NV-Embed-v2 as the dense retriever to ensure a fair and consistent comparison across different systems. 
For downstream reasoning and answer generation, we use Llama-3.3-70B-Instruct~\citep{Grattafiori2024llama} as the backbone LLM.
To evaluate the generality and plug-in nature of \method{}, we integrate it into two representative and influential graph-based RAG systems, i.e., GraphRAG and HippoRAG 2, and adopt NV-Embed-v2 as the dense retriever of the vanilla RAG system. 
In both cases, \method{} is applied as a pre-retrieval adapter without modifying the original components of the underlying systems.

For some baselines, we report the results directly from HippoRAG 2 when the experimental settings are identical.
For methods involved in adapter integration, including GraphRAG, HippoRAG 2, and vanilla RAG with NV-Embed-v2, we re-implement and re-evaluate them under our unified setup for a fair comparison.
Additional implementation details, including the dataset statistics and router training details, are provided in Appx.~\S\ref{appendix:Implementation Details}.
\begin{table*}[htbp]
\centering
\caption{Retrieval performance comparison in terms of Recall@2 and Recall@5. GraphRAG and LightRAG are not compared as they do not conduct passage retrieval.}
\small
\renewcommand{\arraystretch}{1.1}
\resizebox{.9\textwidth}{!}{%
\begin{tabular}{lcccccccc}
\toprule
\multirow{2}{*}{\textbf{Method}}
 & \multicolumn{2}{c}{\textbf{HotpotQA}} 
 & \multicolumn{2}{c}{\textbf{2Wiki}} 
 & \multicolumn{2}{c}{\textbf{MuSiQue}}
 & \multicolumn{2}{c}{\textbf{Average}}\\
\cmidrule(lr){2-3} \cmidrule(lr){4-5} \cmidrule(lr){6-7} \cmidrule(lr){8-9}
 & \textbf{R@2} & \textbf{R@5} 
 & \textbf{R@2} & \textbf{R@5}
 & \textbf{R@2} & \textbf{R@5}
 & \textbf{R@2} & \textbf{R@5}\\
\midrule
\rowcolor{gray!10}
\multicolumn{9}{c}{\textit{\textbf{Vanilla RAG}}}\\
\midrule
BM25~\citep{Robertson1994some} & 57.3 & 74.8 & 55.3 & 65.3 & 32.4 & 43.5 & 48.3 & 61.2 \\
Contriever~\citep{Izacard2022unsupervised} & 58.4 & 75.3 & 46.6 & 57.5 & 34.8 & 46.6 & 46.6 & 59.8 \\
NV-Embed-v2~\citep{Lee2025nvembed} & \underline{84.1} & \underline{94.4} & 69.1 & 76.6 & 52.7 & 69.4 & 68.6 & 80.1 \\
\midrule
\rowcolor{gray!10}
\multicolumn{9}{c}{\textit{\textbf{Graph-based RAG}}}\\
\midrule
RAPTOR~\citep{Sarthi2024raptor}  & 76.8 & 86.9 & 58.3 & 66.2 & 47.0 & 57.8 & 60.7 & 70.3 \\
HippoRAG~\citep{Gutierrez2024hipporag}  & 60.4 & 77.3 & 71.9 & \underline{90.4} & 41.2 & 53.2 & 57.8 & 73.6 \\
HippoRAG 2~\citep{Gutierrez2025from} & 83.3 & \textbf{96.1} & \underline{76.5} & \textbf{91.2} & \textbf{54.5} & \textbf{73.3} & \underline{71.4} & \textbf{86.9} \\
\rowcolor{lavender}
\textbf{\method{} (HippoRAG 2)} & \textbf{85.9} & \textbf{96.1} & \textbf{77.0} & 85.1 & \underline{54.2} & \underline{72.8} & \textbf{72.4} & \underline{84.7} \\
\bottomrule
\end{tabular}}
\label{tab:main_results_retrieval}
\end{table*}
\begin{table*}[t]
\centering
\caption{QA performance comparison with different routers when plugged into GraphRAG and HippoRAG 2.}
\small
\renewcommand{\arraystretch}{1.1}
\resizebox{.75\textwidth}{!}{%
\begin{tabular}{lcccccccccccc}
\toprule
\multirow{2}{*}{\textbf{Method}}
 & \multicolumn{2}{c}{\textbf{HotpotQA}} 
 & \multicolumn{2}{c}{\textbf{2Wiki}} 
 & \multicolumn{2}{c}{\textbf{MuSiQue}}
 & \multicolumn{2}{c}{\textbf{Average}}\\
\cmidrule(lr){2-3} \cmidrule(lr){4-5} \cmidrule(lr){6-7} \cmidrule(lr){8-9}
 & \textbf{EM} & \textbf{F1} 
 & \textbf{EM} & \textbf{F1}
 & \textbf{EM} & \textbf{F1}
 & \textbf{EM} & \textbf{F1}\\
\midrule
\rowcolor{gray!10}
\multicolumn{9}{c}{\textit{\textbf{GraphRAG}}}\\
\midrule
\textit{w/} NER-Router & 57.5 & 71.5 & \underline{49.7} & \underline{54.7} & \textbf{29.0} & \underline{39.3} & 45.4 & \underline{55.2} \\
\textit{w/} LLM-Router & \textbf{59.7} & \underline{72.6} & 49.3 & 54.1 & 27.7 & 38.2 & \underline{45.6} & 55.0 \\
\rowcolor{lavender}
\textit{w/} \method{}-Router & \underline{58.8} & \textbf{73.0} & \textbf{55.5} & \textbf{62.4} & \underline{28.5} & \textbf{40.8} & \textbf{47.6} & \textbf{58.7} \\
\midrule
\rowcolor{gray!10}
\multicolumn{9}{c}{\textit{\textbf{HippoRAG 2}}}\\
\midrule
\textit{w/} NER-Router & \underline{62.0} & \underline{74.7} & \underline{59.6} & \underline{64.6} & \underline{35.4} & \underline{46.2} & \underline{52.3} & \underline{61.8} \\
\textit{w/} LLM-Router & 61.4 & 74.6 & 59.4 & 64.2 & 35.1 & \underline{46.2} & 52.0 & 61.7 \\
\rowcolor{lavender}
\textit{w/} \method{}-Router & \textbf{62.5} & \textbf{75.8} & \textbf{64.8} & \textbf{71.7} & \textbf{38.1} & \textbf{49.6} & \textbf{55.1} & \textbf{65.7} \\
\bottomrule
\end{tabular}}
\label{tab:router}
\end{table*}

\subsection{Main Results (RQ1)}
We conduct comprehensive evaluations on three multi-hop QA benchmarks to assess the effectiveness of \method{}. \Cref{tab:main_results_QA} and~\Cref{tab:main_results_retrieval} report the QA and retrieval results, respectively. From these results, we make the following observations:

\textbf{(1) \method{} reduces reliance on graph-based RAG while maintaining comparable QA performance.}
When integrated with HippoRAG 2, \method{} achieves the best average QA performance across all methods, slightly surpassing the original method. When applied to GraphRAG, a clear improvement over GraphRAG itself is observed, demonstrating that \method{} does not compromise answer accuracy while routing approximately 45\% of queries to vanilla RAG on average across datasets.

\textbf{(2) \method{} generalizes well across different graph-based RAG systems and unseen settings.}
Although the router is trained using supervision derived from HotpotQA and HippoRAG 2, consistent performance trends are observed when \method{} is plugged into both HippoRAG 2 and GraphRAG, despite their substantially different graph construction and reasoning paradigms, as well as across unseen QA benchmarks.

\textbf{(3) \method{} maintains strong retrieval effectiveness while selectively reducing graph retrieval usage.}
As shown in \Cref{tab:main_results_retrieval}, integrating \method{} with HippoRAG 2 slightly improves Recall@2 and yields comparable Recall@5 on average. These results demonstrate that \method{} achieves a favorable balance between retrieval cost and retrieval effectiveness by selectively engaging graph-based retrieval only when it is likely to be beneficial.

\begin{figure}[ht]
    \centering
    \includegraphics[width=\linewidth]{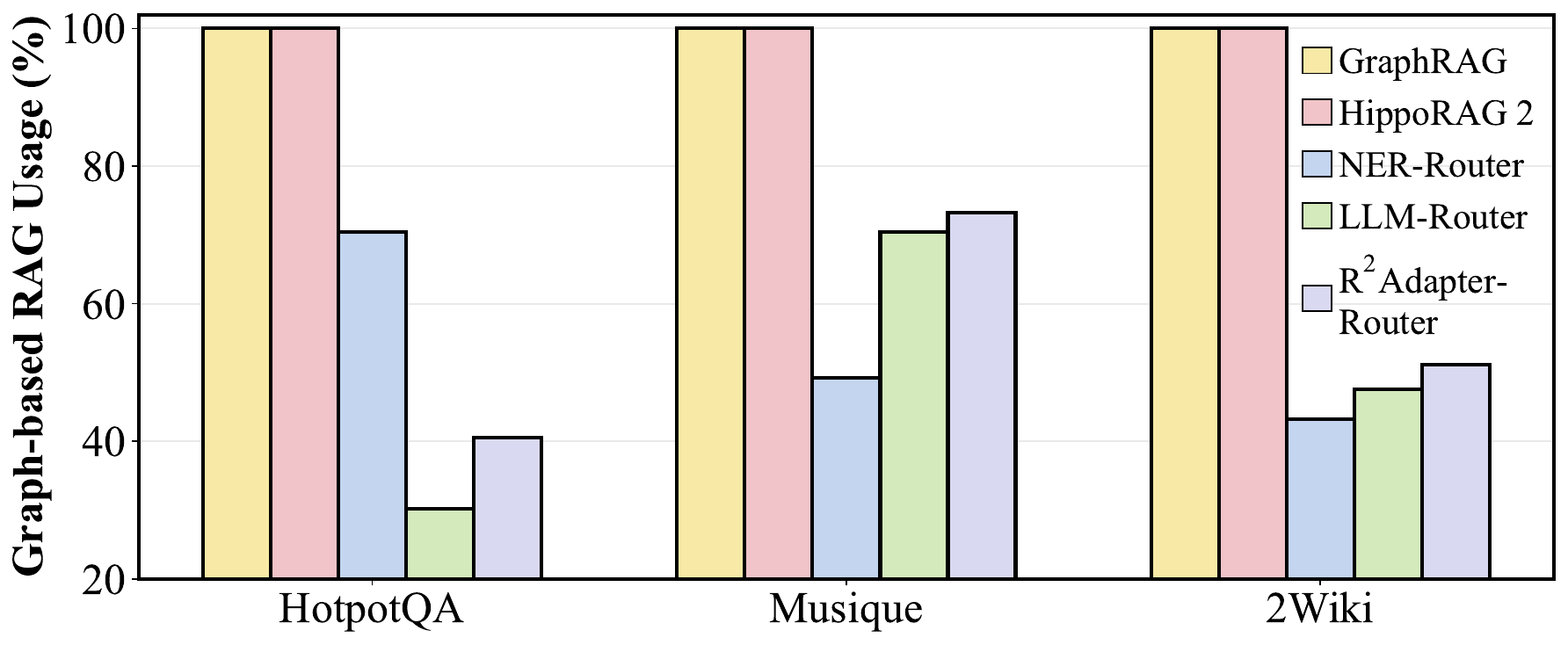}
    \caption{The proportion of queries routed to graph-based RAG by different methods.}
    \label{fig:graphrag usage}
\end{figure}

\subsection{Analysis of the Router (RQ2)}
To analyse the effectiveness of the router and the efficiency gain it brings, we compare our router with two routing baselines, namely NER-Router and LLM-Router. In this analysis, \method{}-Router denotes the adapter equipped with the router only, without query rewriting.
As shown in~\Cref{tab:router}, our router consistently yields better or comparable QA performance than both NER- and LLM-based routers when integrated with GraphRAG and HippoRAG 2. In contrast, NER-Router and LLM-Router exhibit unstable behavior across datasets, reflecting their limited ability to accurately capture when graph-based RAG is truly beneficial. These results indicate that the router of \method{} provides a more reliable routing signal.

We further examine routing efficiency by measuring the proportion of queries routed to the graph-based RAG. As illustrated in~\Cref{fig:graphrag usage}, \method{}-Router substantially reduces graph usage compared to always-on graph-based RAG while maintaining competitive QA performance, with the largest reduction observed on HotpotQA (59\%). Moreover, under a similar average graph routing ratio to NER- and LLM-based routers, it achieves markedly stronger answer accuracy.

\begin{figure*}[th]
    \centering
    \includegraphics[width=\linewidth]{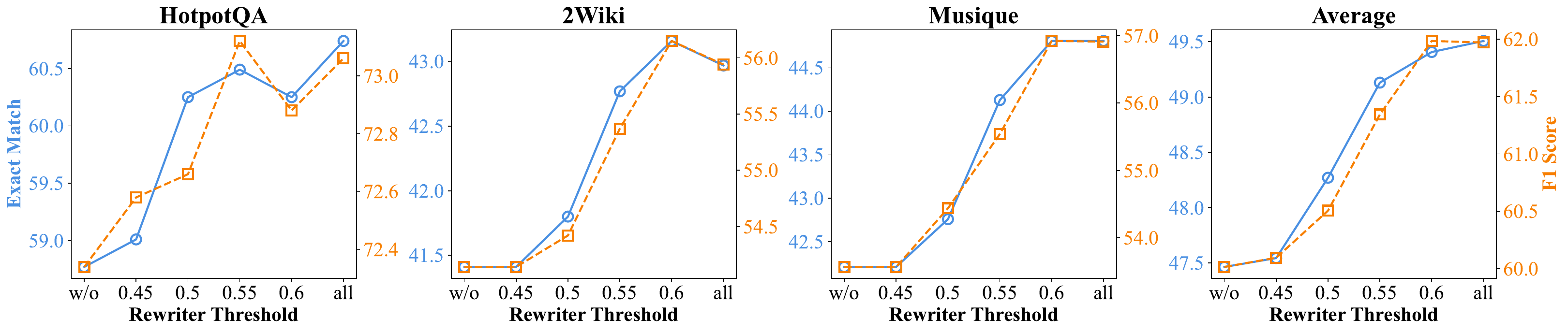}
    \caption{Effect of the rewriter threshold on answer accuracy across datasets (integrated with HippoRAG 2).}
    \label{fig:rewriter thresholds}
\end{figure*}

\subsection{Analysis of the Rewriter (RQ3)}
We further analyse the effect of the rewriter by varying the threshold $\epsilon$ in~\Cref{equ:rewriter threshold}, which controls whether a query routed to the graph-based RAG system is rewritten. Since query rewriting is only applied after routing to the graph-based RAG, all results in this section are reported on this routed subset.
\revise{As shown in~\Cref{fig:rewriter thresholds}}, introducing query rewriting consistently improves answer accuracy compared to the setting without rewriting. As $\epsilon$ increases, more routed queries are rewritten, leading to steady gains in both EM (blue) and F1 (yellow) across datasets. This suggests that rewriting helps better align complex queries with graph-based RAG.
However, the performance gains saturate when rewriting is applied to all routed queries, indicating that selective rewriting is sufficient and avoids unnecessary perturbations to already well-formed queries.
To provide a more intuitive understanding of how query rewriting affects graph-based reasoning, we present a representative case study in \Cref{tab:rewriter_case}, highlighting how query rewriting helps disambiguate implicit relations by explicitly decomposing the original question into structured sub-queries. 

\begin{table}[t]
\centering
\caption{Case study comparing answers produced by HippoRAG 2 before and after plugging in \method{}.}
\small
\renewcommand{\arraystretch}{1.15}
\begin{tabular}{p{2cm} p{4.8cm}}
\toprule
\textbf{Method} & \textbf{Answer} \\
\midrule
HippoRAG 2 & Maryland \textcolor{red!70!black}{(\textit{incorrect})} \\
\multirow{2}{*}{\textit{w/} \method{}} & from the port of Baltimore west to Sandy Hook \textcolor{green!60!black}{(\textit{correct})}\\
\bottomrule
\end{tabular}

\vspace{0.3em}

\begin{tabular}{p{0.3cm} p{6.5cm}}
\toprule
\multirow{2}{*}{\textbf{$q$}} & Where was the original line of the railroad William Howard worked for? \\
\midrule
\multirow{2}{*}{\textbf{$q_r$}} & \texttt{[William\_Howard, worked\_for, <railroad>]} \\
 & \texttt{[<railroad>, had\_original\_line, ?]} \\
\bottomrule
\end{tabular}
\label{tab:rewriter_case}
\end{table}

\section{Conclusions}
In this paper, we present \method{}, a light routing and rewriting adapter for efficient hybrid RAG systems. Motivated by the observation that real-world queries exhibit diverse reasoning requirements, \method{} dynamically allocates queries between vanilla and graph-based RAG. By routing only queries that genuinely benefit from graph-based reasoning and selectively rewriting uncertain queries, \method{} achieves a practical balance between inference cost and reasoning capability.
Extensive experiments on multi-hop QA benchmarks demonstrate that \method{} substantially reduces graph retrieval usage while maintaining comparable answer accuracy across mainstream graph-based RAG methods. As a model-agnostic plug-in that avoids costly LLM-based query classification, \method{} offers a practical and scalable solution for hybrid text-graph RAG systems.

\section*{Limitations}
While \method{} effectively balances efficiency and accuracy, it possesses several limitations. The overall performance of the pipeline is inherently bound by the vanilla and graph-based RAG systems \method{} plugs into. In particular, the effectiveness of the pipeline relies heavily on the quality of the dense retrievers and underlying LLMs, as any limitations or biases in these models can propagate through the system, affecting both retrieval coverage and answer accuracy. The router also has intrinsic constraints as it is trained on automatically constructed data under a specific dense retriever. If the retriever is replaced, the router may need retraining to fit the new retrieval distribution and maintain optimal alignment, though it still provides reasonable routing. Also, the rewriter employs a static threshold-based policy rather than an adaptive learning mechanism, which may not fully capture the complex characteristics of queries that would benefit most from rewriting. Therefore, future work could explore more sophisticated or learned rewriting strategies to adaptively select which queries to rewrite.


\bibliography{custom}

\clearpage
\appendix

\section{Evaluation Datasets and Baselines}
\subsection{Evaluation Datasets}
The statistics of datasets, including the details of the graph constructed by HippoRAG 2 and GraphRAG, are shown in~\Cref{tab:dataset_statistics}.
\begin{table*}[htbp]
\centering
\caption{Dataset statistics.}
\label{tab:dataset_statistics}
\begin{tabular}{lccc}
\toprule
 & \textbf{HotpotQA} & \textbf{2Wiki} & \textbf{MuSiQue} \\
\midrule
\# of queries  & 1,000 & 1,000 & 1,000 \\
\# of passages & 9,811 & 6,119 & 11,656 \\
\# of nodes (HippoRAG 2)  & 72,587 & 40,805 & 76,719 \\
\# of edges (HippoRAG 2) & 115,778 & 62,806 & 125,769 \\
\# of entities (GraphRAG) & 59,564 & 33,874 & 62,385 \\
\# of communities (GraphRAG) & 13,306 & 7,590 & 14,370 \\
\# of relationships (GraphRAG) & 109,947 & 57,786 & 122,395 \\
\bottomrule
\end{tabular}
\end{table*}

\begin{table*}[htbp]
\caption{LLM-Router prompt template for query classification.}
\label{tab:prompt_template_LLM_router}
\centering
\begin{tcolorbox}[
    colback=black!5, 
    colframe=black!70!white, 
    title=\textbf{LLM-Router Prompt Template},
    fonttitle=\bfseries,
    arc=3mm, 
    boxrule=0.8pt 
]
\begin{tabularx}{\textwidth}{X}
You are an AI model tasked with classifying queries into one of two categories based on their complexity and reasoning requirements.\\

\par\bigskip
Category Definitions

\par\bigskip
1. Fact-Based Queries\\
- The answer can be directly retrieved from a knowledge source or requires details. \\
- The query does not require multi-step reasoning, inference, or cross-referencing multiple sources.\\

\par\bigskip
2. Reasoning-Based Queries \\
- The answer cannot be found in a single lookup and requires cross-referencing multiple sources, logical inference, or multistep reasoning.\\

\par\bigskip
Output Rules: \\
- Output ONLY a single number.\\
- Output 0 if the query is Fact-Based.\\
- Output 1 if the query is Reasoning-Based.\\
- Do NOT include any explanation or extra text.\\

\par\bigskip
Query: \{\texttt{user query}\}
\end{tabularx}
\end{tcolorbox}
\end{table*}

\subsection{Baselines}
\textbf{Vanilla RAG.}
Vanilla RAG systems are equipped with different retrievers:
\begin{itemize}
    \item \textbf{BM25}~\citep{Robertson1994some}, a classical sparse retriever that ranks documents based on term frequency and inverse document frequency.
    \item \textbf{Contriever}~\citep{Izacard2022unsupervised}, a dense retriever trained via unsupervised contrastive learning.
    \item \textbf{NV-Embed-v2}~\citep{Lee2025nvembed}, a recent strong dense retriever that trains LLM as a versatile embedding model.
\end{itemize}

\smallskip
\noindent \textbf{Graph-based RAG.}
\begin{itemize}
    \item \textbf{GraphRAG}~\citep{Edge2024local}, a RAG system that builds a community-based knowledge graph to enable comprehensive global summarization and local entity retrieval.
    \item \textbf{LightRAG}~\citep{Guo2025lightrag}, a dual-level RAG system that integrates graph structures to capture both low-level entity details and high-level global topics.
    \item \textbf{RAPTOR}~\citep{Sarthi2024raptor}, a tree-based RAG system that recursively clusters and summarizes text chunks to provide a hierarchical representation of documents across different levels of abstraction.
    \item \textbf{HippoRAG}~\citep{Gutierrez2024hipporag}, a RAG system inspired by hippocampal memory that mimics human-like associative memory through personalized PageRank over a collaborative knowledge graph.
    \item \textbf{HippoRAG 2}~\citep{Gutierrez2025from}, an advanced iteration of HippoRAG that introduces passage nodes to support more complex, multi-hop reasoning.
\end{itemize}

\smallskip
\noindent \textbf{Routing Baselines.}
\begin{itemize}
    \item \textbf{NER-Router} routes queries according to the number of named entities identified by spaCy~\citep{Honnibal2020spaCy}, inspired by the heuristic routing strategy in HybGRAG~\citep{Lee2025hybgrag}, while the latter one adopts an LLM to extract named entities and relations. Specifically, if fewer than two named entities are detected in a query, it is routed to vanilla RAG; otherwise, it is routed to the graph-based RAG. This baseline reflects a lightweight, rule-based assumption that queries involving multiple entities are more likely to benefit from structured, graph-based reasoning.
    \item \textbf{LLM-Router}~\citep{Han2025rag} employs an LLM to classify each query as either fact-based or reasoning-based. Fact-based queries are routed to vanilla RAG, while reasoning-based queries are routed to graph-based RAG. The classification is performed using a fixed prompt template, which is detailed in~\Cref{tab:prompt_template_LLM_router}. This baseline represents a more expressive but computationally expensive routing strategy that relies on LLM-based semantic understanding.
\end{itemize}

\begin{table*}[htbp]
\centering
\caption{Detailed statistics of the router training corpus. Abbreviations V., G., Pref., and Suit. denote Vanilla, Graph-based, Preference, and Suitability, respectively.}
\label{tab:dataset_statistics}
\begin{tabular}{lcccc}
\toprule
\textbf{Split} & \textbf{\# Samples} & \textbf{V. RAG Pref.} & \textbf{G. RAG Pref.} & \textbf{G. RAG Suit.} \\
\midrule
Training   & 11,975 & 10,508 & 1,467 & 9,804 \\
Validation & 1,496  & 1,332 & 164 & 1,236 \\
Testing    & 1,498 & 1,305 & 193 & 1,221 \\
\midrule
\textbf{Total} & 14,969 & 13,145 & 1,824 & 12,261 \\
\bottomrule
\end{tabular}
\end{table*}
\begin{table*}[htbp]
\centering
\caption{Training corpus statistics by question type.}
\label{tab:qtype_stats}
\begin{tabular}{lccc}
\toprule
\textbf{Type} & \textbf{\# Samples} & \textbf{G. RAG Pref.} & \textbf{G. RAG Suit.} \\
\midrule
\texttt{BRIDGE}      & 11,761 & 15.4\% & 93.9\% \\
\texttt{COMPARISON}  & 3,208  & 0.2\%  & 94.6\% \\
\bottomrule
\end{tabular}
\end{table*}
\begin{table*}[htbp]
\centering
\caption{NER statistics of the training corpus across different routing preferences.}
\label{tab:ner_stats}
\begin{tabular}{lccc}
\toprule
\textbf{Metric} & \textbf{Overall} & \textbf{V. RAG Pref.} & \textbf{G. RAG Pref.} \\
\midrule
Avg. Entities / Sample & 2.26 & \textbf{2.30} & 1.97 \\
Samples w/ Entities    & 96.5\% & \textbf{96.8\%} & 94.2\% \\
\bottomrule
\end{tabular}
\end{table*}

\begin{table*}[htbp]
\centering
\caption{Entity type distribution by routing preference. Only top categories are shown.}
\label{tab:entity_type_dist}
\begin{tabular}{lccc}
\toprule
\textbf{Entity Type} & \textbf{V. RAG Pref.} & \textbf{G. RAG Pref.} & \textbf{$\Delta$} \\
\midrule
\texttt{PERSON} & 30.8\% & 23.9\% & \textcolor{red!70!black}{-6.9\%} \\
\texttt{ORG}    & 18.4\% & 18.7\% & \textcolor{green!60!black}{+0.3\%} \\
\texttt{DATE}   & 11.2\% & 16.0\% & \textcolor{green!60!black}{\textbf{+4.8\%}} \\
\texttt{GPE}    & 12.1\% & 10.4\% & \textcolor{red!70!black}{-1.7\%} \\
\texttt{NORP}   & 9.2\%  & 10.1\% & \textcolor{green!60!black}{+0.9\%} \\
\bottomrule
\end{tabular}
\end{table*}

\begin{algorithm*}[t]
\caption{Router Training Process}
\label{alg:router_training}
\KwIn{Training set $\mathcal{D}_{train}$, Validation set $\mathcal{D}_{val}$, Pre-trained backbone model $M_{pre}$, Learning rate $\eta$, Epochs $E$, Balancing factor $\alpha$, Focusing parameter $\gamma$, Evaluation frequency $N_{eval}$}
\KwOut{Optimal model $M^*$, Best routing threshold $\tau^*$}

$M \gets M_{pre}$\;
$S_{best} \gets 0, \tau^* \gets 0.0, step \gets 0$\;

\For{epoch $e = 1$ \KwTo $E$}{
    \ForEach{batch $\mathcal{B} \subset \mathcal{D}_{train}$}{
        \tcc{Forward pass and score prediction}
        $[\hat{p}_p, \hat{p}_g] \gets M(q), \forall q \in \mathcal{B}$ \tcp*{Compute suitability probabilities}
        \tcc{Independent Focal Loss optimization}
        $\mathcal{L}_{router} \gets \frac{1}{2} \sum_{k \in \{p, g\}} \mathcal{L}_k(\hat{p}_k, y_k; \alpha, \gamma)$ \tcp*{Calculate dual-output loss}
        Update $M$ weights via $\text{AdamW}(\mathcal{L}_{router}, \eta)$ \tcp*{Perform optimization}
        
        $step \gets step + 1$\;
        
        \tcc{Periodic threshold optimization on validation set}
        \If{$step \pmod{N_{eval}} == 0$}{
            Predict $\hat{p}_p, \hat{p}_g$ for all samples in $\mathcal{D}_{val}$ using $M$\;
            \For{$\tau \in [0.0, 0.5]$ with step $\Delta \tau$}{
                $preds \gets (\hat{p}_g - \hat{p}_p > \tau)$ \tcp*{Preference decision}
                $Score(\tau) \gets \frac{1}{2} \left( \text{F1}(preds) + \text{Recall}(preds) \right)$\;
                \If{$Score(\tau) > S_{best}$}{
                    $S_{best} \gets Score(\tau)$\;
                    $\tau^* \gets \tau$\;
                    $M^* \gets M$ \tcp*{Save the best performing model state}
                }
            }
        }
    }
}
\Return $M^*, \tau^*$
\end{algorithm*}

\section{Implementation Details}
\label{appendix:Implementation Details}
\subsection{Router Training Corpus}
The router is trained on the automatically labeled corpus constructed in~\S\ref{sec:Training Corpus Construction}. We randomly partition the dataset into training, validation, and testing sets following an 8:1:1 ratio. The detailed statistics of the corpus and the distribution of question types are presented in~\Cref{tab:dataset_statistics} and~\Cref{tab:qtype_stats}, respectively.
By conducting a granular analysis of this corpus across various dimensions, we aim to identify the specific scenarios where graph-based RAG holds a decisive advantage over vanilla RAG. These findings provide empirical guidelines for designing more effective and cost-efficient routing strategies in complex RAG systems.

\smallskip
\noindent \textbf{Question Type Preference.} 
As shown in~\Cref{tab:qtype_stats}, graph-based RAG exhibits a distinct advantage in handling \texttt{BRIDGE} questions compared to \texttt{COMPARISON} questions. Specifically, 15.4\% of \texttt{BRIDGE} queries are explicitly preferred to be routed to the graph paradigm. In contrast, \texttt{COMPARISON} queries are almost exclusively preferred by vanilla RAG (99.8\%), despite the high suitability of the graph paradigm (94.6\%). This suggests that while the graph structure is capable of representing comparison relations, its structural traversal is particularly indispensable for bridging disjoint reasoning paths.

To further illustrate the distinction between routing preferences, we provide two representative instances from our training corpus.
Listing~\ref{lst:case_comparison} shows a \texttt{COMPARISON} query that both the vanilla and graph-based RAG systems achieve identical retrieval performance with an average rank of 1.5. While the query is marked as ``graph-acceptable'', the final preference is assigned to passage-based vanilla RAG, as the paradigm with lower computational complexity and higher inference throughput is preferred when capabilities are comparable.
Listing~\ref{lst:case_bridge} illustrates a \texttt{BRIDGE} query requiring a multi-hop traversal from a specific creative work (\texttt{Oracular Spectacular}) to its creator, and finally to the lead singer. Here, the graph-based RAG system achieves a superior average rank of 1.5, whereas the vanilla RAG system lags at 2.5. Consequently, the graph-based RAG is preferred.

\begin{lstlisting}[caption={Example of a \texttt{COMPARISON}-type query.}, label={lst:case_comparison}, language=json]
{
  "query": "What prestigious award have Bertrand Russell and Gunter Grass won?",
  "type": "comparison",
  "preference": "passage",
  "acceptable": "graph",
  "passage_avg_rank": 1.5,
  "graph_avg_rank": 1.5,
  "ner_entities": [
    {
      "text": "Bertrand Russell", 
      "label": "PERSON"
    },
    {
      "text": "Gunter Grass", 
      "label": "PERSON"
    }
  ]
}
\end{lstlisting}

\begin{lstlisting}[caption={Example of a \texttt{BRIDGE}-type query.}, label={lst:case_bridge}, language=json]
{
  "query": "Who is the lead singer of a rock band that created an album called 'Oracular Spectacular'?",
  "type": "bridge",
  "preference": "graph",
  "acceptable": "",
  "passage_avg_rank": 2.5,
  "graph_avg_rank": 1.5,
  "ner_entities": [
    {
      "text": "Oracular Spectacular", 
      "label": "WORK_OF_ART"
    }
  ]
}
\end{lstlisting}

\begin{center}
\fbox{\parbox{0.95\linewidth}{\textit{\textbf{Takeaway 1.} Graph-based RAG is more specialized for bridge-type queries, where explicit multi-hop path traversal is required, while comparison-type queries are more efficiently handled by passage-level synthesis.}}}
\end{center}

\smallskip
\noindent \textbf{NER Feature Analysis.} 
We further investigate the relationship between named entities extracted by spaCy and routing preference. As shown in \Cref{tab:ner_stats}, contrary to the heuristic intuition that more entities necessitate graph-based retrieval, our statistics reveal that vanilla RAG-preferred queries actually contain a higher average number of entities per sample (2.30) than graph-based RAG-preferred ones (1.97). Furthermore, the entity presence rate is also slightly higher in the vanilla RAG group (96.8\% vs. 94.2\%). This indicates that the necessity of graph-based RAG is driven by the relational complexity and reasoning depth between entities, rather than the mere quantity of entities mentioned in the query.

\begin{center}
\fbox{\parbox{0.95\linewidth}{\textit{\textbf{Takeaway 2.} Entity count is not a reliable proxy for routing decisions. The complexity of reasoning chains outweighs surface-level entity density in determining the suitability of graph-based RAG.}}}
\end{center}

Beyond entity counts, the distribution of entity types, as shown in \Cref{tab:entity_type_dist}, reveals significant qualitative differences. While \texttt{PERSON} and \texttt{ORG} remain the dominant types in both groups, queries with a graph-based RAG preference exhibit a substantially higher proportion of \texttt{DATE} entities (16.0\%) compared to those with a vanilla RAG preference (11.2\%). This discrepancy suggests that graph-based RAG is frequently invoked for queries involving temporal constraints and reasoning. The structured nature of graph representations is likely more effective at capturing the relationships implied by such temporal entities than flat passage-level text embeddings.

\begin{center}
\fbox{\parbox{0.95\linewidth}{\textit{\textbf{Takeaway 3.} Graph-based RAG may possess a superior potential for handling temporal and chronologically constrained reasoning tasks compared to vanilla RAG.}}}
\end{center}

\subsection{Router Training Details}
We provide the comprehensive router training and threshold optimization procedure in~\Cref{alg:router_training}. The router is initialized with a pre-trained DeBERTa-v3-base~\citep{He2023deberta} backbone $M_{pre}$ and fine-tuned on our constructed training corpus $\mathcal{D}_{train}$. 
Statistics in~\Cref{tab:dataset_statistics} exhibit a significant imbalance in preference labels, where vanilla RAG preferences dominate the distribution. This imbalance, coupled with the presence of challenging ``boundary'' samples, complicates direct preference learning. Specifically, for queries where both vanilla and graph-based RAG yield acceptable answers, vanilla RAG is marked as the preferred choice due to its higher efficiency. Learning such efficiency-driven preferences introduces additional difficulty to the model, as it struggles to distinguish them from capability-driven requirements. To mitigate this, we transform the learning objective from direct preference classification to multidimensional suitability estimation. By learning suitability features for each paradigm independently, the router more effectively captures the underlying applicability of each method, particularly for these overlapping cases.

During each training step, the model estimates the suitability scores $[\hat{p}_p, \hat{p}_g]$ for vanilla and graph-based RAG, respectively. These predictions are supervised by the independent focal loss $\mathcal{L}_{router}$, which effectively addresses class imbalance and emphasizes hard examples by down-weighting easy negatives.
To ensure the router generalizes well under the asymmetric risks of misrouting, we perform a periodic threshold search on the validation set $\mathcal{D}_{val}$ every $N_{eval}$ steps. Instead of using a fixed classification threshold, we iterate through candidate thresholds $\tau$ within the range $[0.0, 0.5]$ with a step size of $\Delta \tau$. For each candidate, we calculate a balanced metric $\mathrm{Score}(\tau)$ that jointly considers the F1 score and the recall of the graph-based class, reflecting our priority on not missing queries that necessitate graph-based reasoning. The model state $M^*$ and the threshold $\tau^*$ that achieve the highest validation score are preserved for inference.

All hyperparameters used in this process are summarized in~\Cref{tab:Hyperparameters}. The model is optimized using the AdamW optimizer for a single epoch, with a linear learning rate scheduler and a warmup phase covering the first 10\% of total training steps.
\begin{table}[t]
    \centering
    \caption{Hyperparameters for router training.}
    \label{tab:Hyperparameters}
    \begin{tabular}{cc}
    \toprule
        \textbf{Hyperparameter} & \textbf{Value} \\
    \midrule
        Learning Rate & $1 \times 10^{-5}$ \\
        Batch Size & 64 \\
        Training Epochs & 1 \\
        Warmup Ratio & 0.1 \\
        Evaluation Steps & 50 \\
        $\alpha$ & 0.25 \\
        $\gamma$ & 2.0 \\
        $\Delta \tau$ & 0.02 \\
    \bottomrule
    \end{tabular}
\end{table}

\subsection{Rewriter Prompt Template}
\label{appendix: Rewriter Prompt Template}
The prompt template of the rewriter is shown in~\Cref{tab:prompt_template}.
\begin{table*}[thbp]
\caption{Rewriter prompt template for the query routed to graph-based RAG with low confidence.}
\label{tab:prompt_template}
\centering
\begin{tcolorbox}[
    colback=black!5, 
    colframe=black!70!white, 
    title=\textbf{Rewriter Prompt Template},
    fonttitle=\bfseries,
    arc=3mm, 
    boxrule=0.8pt 
]
\begin{tabularx}{\textwidth}{X}
You are an expert in query rewriting for **knowledge graph-based retrieval systems**.\\
Your task is to convert the given natural language question into one or more **structured triplets** of the form:
\[
\texttt{[head\_entity, relation, tail\_entity]}
\]
Each triplet represents a relational fact or reasoning step used to answer the question.

\par\bigskip
\#\#\# Rules
\begin{enumerate}
    \item The final answer should be represented by **a triplet containing a question mark (?)** in place of the unknown target entity.
   \begin{itemize}
    \item[-] Example: ``Who founded Tesla?'' → \texttt{[?, founded, Tesla\_Inc.]}
    \end{itemize}
    \item Not every triplet must contain a question mark. 
    \begin{itemize}
    \item[-] Intermediate steps may include unknown or inferred entities.
    \end{itemize}
    \item For **unknown intermediate entities**, use **angle brackets `<...>'** with a short descriptive phrase.
    \begin{itemize}
    \item[-] Example: ``Who taught the author of 'The Republic'?'' →
    \item[] \texttt{[<the author of `The Republic'>, wrote, `The Republic']} 
    \item[] \texttt{[?, taught, <the author of `The Republic'>]}
    \end{itemize}
    \item The **order of triplets should follow the reasoning flow** required to answer the question (from known → inferred → target).
    \item Avoid natural language explanations or commentary, only output the final structured reasoning chain.
\end{enumerate}

\#\#\# Output Format

Output as a **list of reasoning steps**, each in the following format:
\[
\texttt{[head\_entity, relation, tail\_entity]}
\]
If multiple steps exist, write them in logical order.

\par\bigskip
Original Question: \{\texttt{user query}\}

\par\bigskip
Triplet Reasoning Chain:
\end{tabularx}
\end{tcolorbox}
\end{table*}

\subsection{Testing Details}
All experiments are conducted on a single node with 8$\times$80GB NVIDIA A100 GPUs. The downstream reasoning and answer generation LLM, Llama-3.3-70B-Instruct~\citep{Grattafiori2024llama}, is deployed using four GPUs with tensor parallelism, while the dense retriever NV-Embed-v2~\citep{Lee2025nvembed} is deployed on a single GPU.

\section{Additional Experiments}
\begin{figure*}[t]
    \centering
    \begin{subfigure}[t]{0.48\linewidth}
        \centering
        \includegraphics[width=\linewidth]{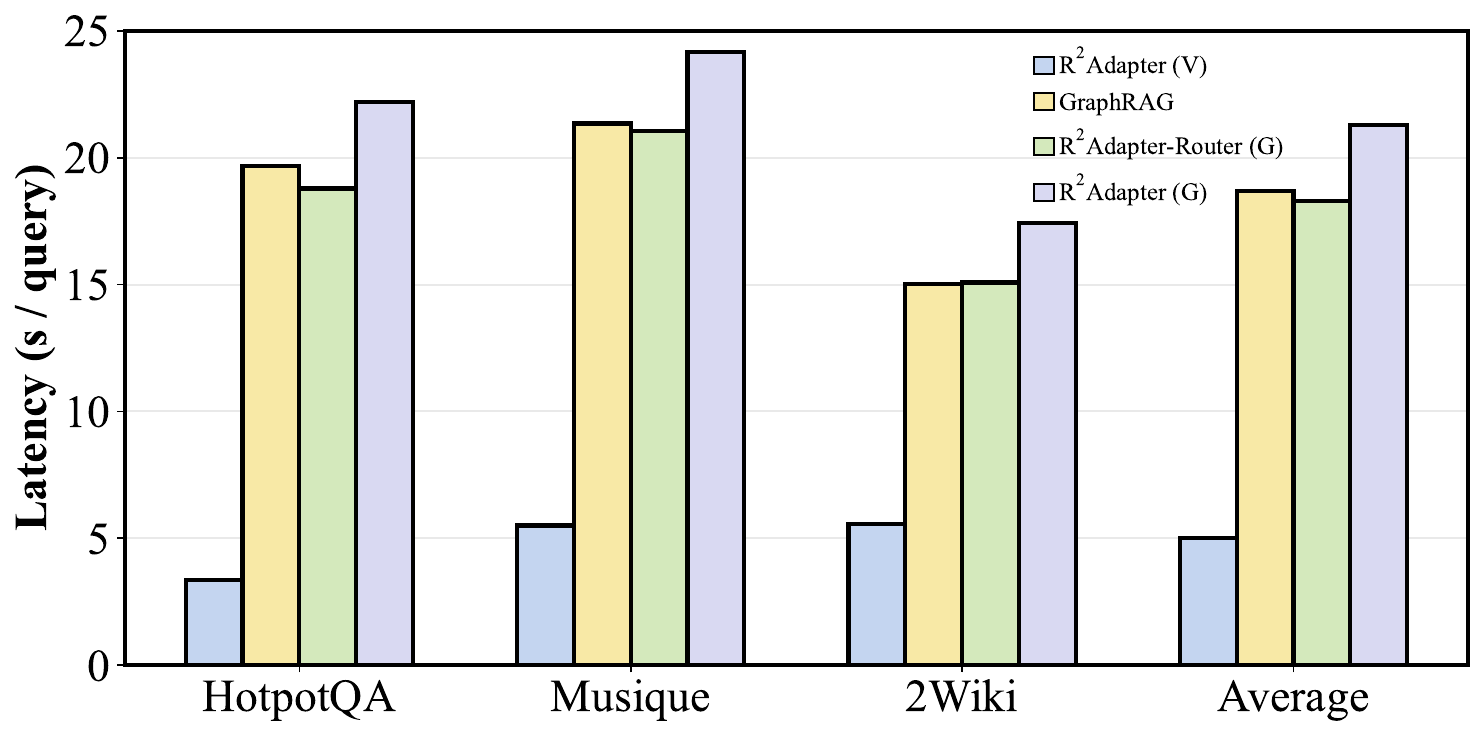}
        \caption{Integrated with GraphRAG}
        \label{fig:graphrag_latency}
    \end{subfigure}
    \hfill
    \begin{subfigure}[t]{0.48\linewidth}
        \centering
        \includegraphics[width=\linewidth]{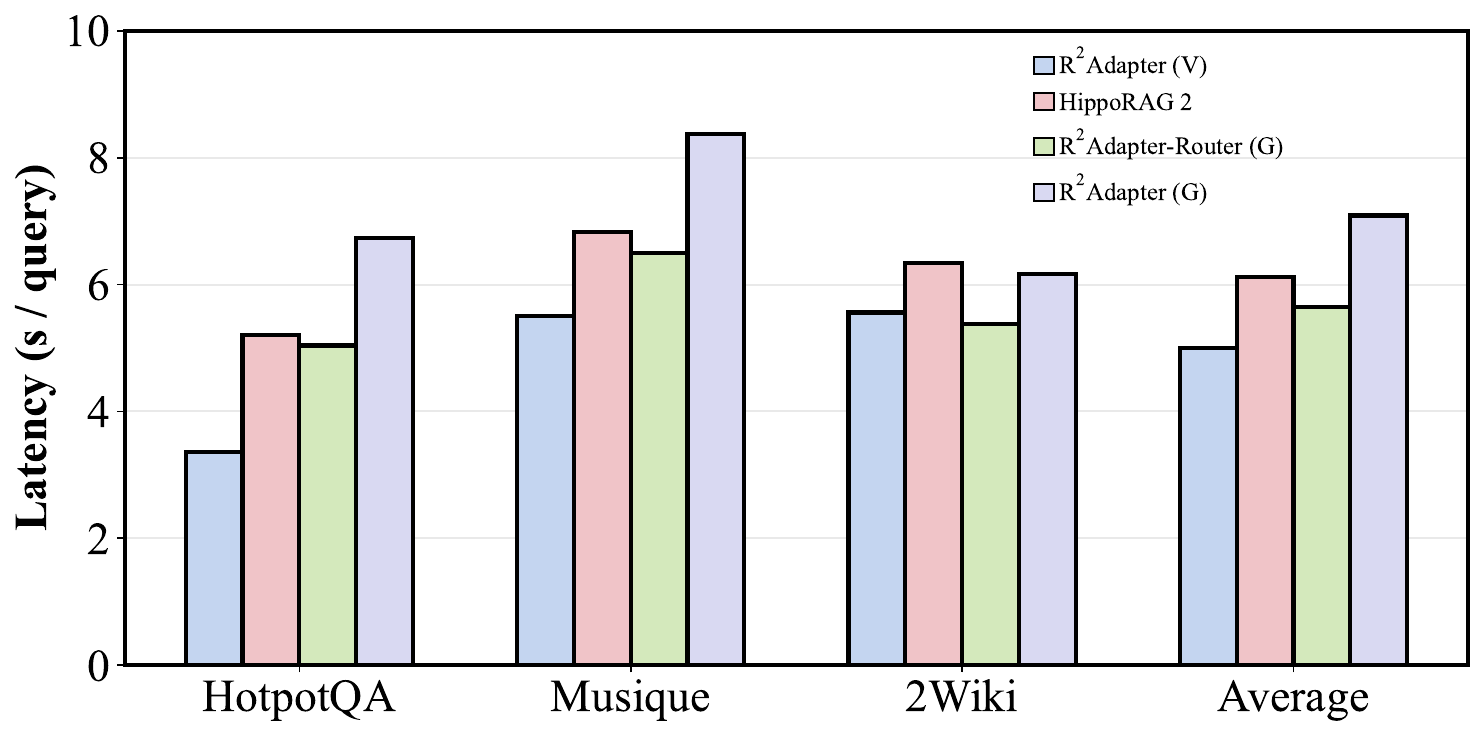}
        \caption{Integrated with HippoRAG 2}
        \label{fig:hipporag_latency}
    \end{subfigure}

    \caption{
    End-to-end latency per query across datasets. \method{}~(V) denotes queries routed to vanilla RAG. \method{}-Router (G) and \method{}~(G) denote queries routed to graph-based RAG using the router-only and adapter-based mechanisms, respectively.
    }
    \label{fig:latency_decomposition}
\end{figure*}
\begin{table}[ht]
\centering
\small
\caption{Average routing time per query and proportion of queries routed to the graph-based RAG with different routers.}
\label{tab:routing time}
\renewcommand{\arraystretch}{1.15}
\begin{tabular}{lcc}
\toprule
\textbf{Method} & \textbf{R. T. (ms)} & \textbf{G. Usage (\%)}\\
\midrule
NER-Router & 7 & 54.27\\
LLM-Router & 74 & 49.40\\
\rowcolor{lavender}
\method{}-Router & 16 & 54.97\\
\bottomrule
\end{tabular}
\end{table}
\begin{figure*}
    \centering
    \includegraphics[width=.7\linewidth]{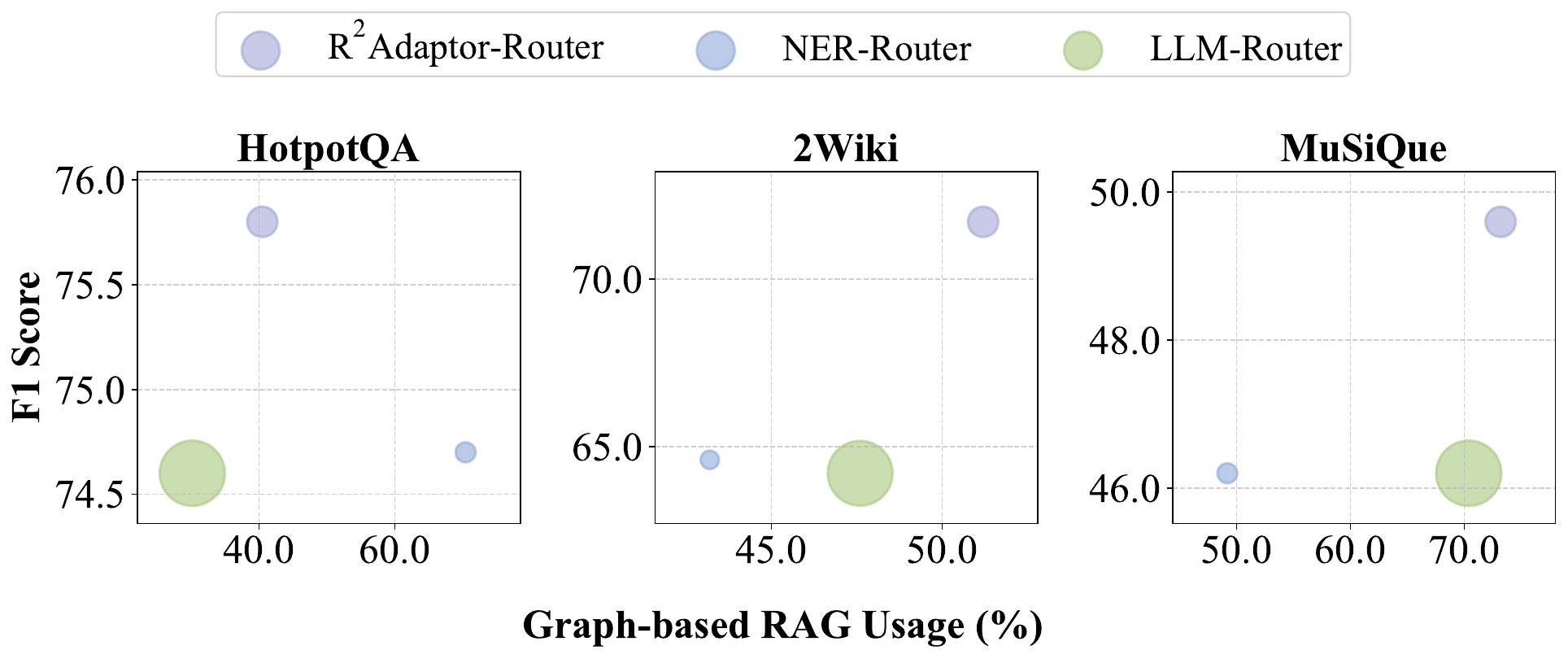}
    \caption{Comparison of routing performance when integrated with HippoRAG 2, while the bubble size indicates routing time per query.}
    \label{fig:router_bubbles}
\end{figure*}
\begin{figure*}[htbp]
    \centering
    \includegraphics[width=\linewidth]{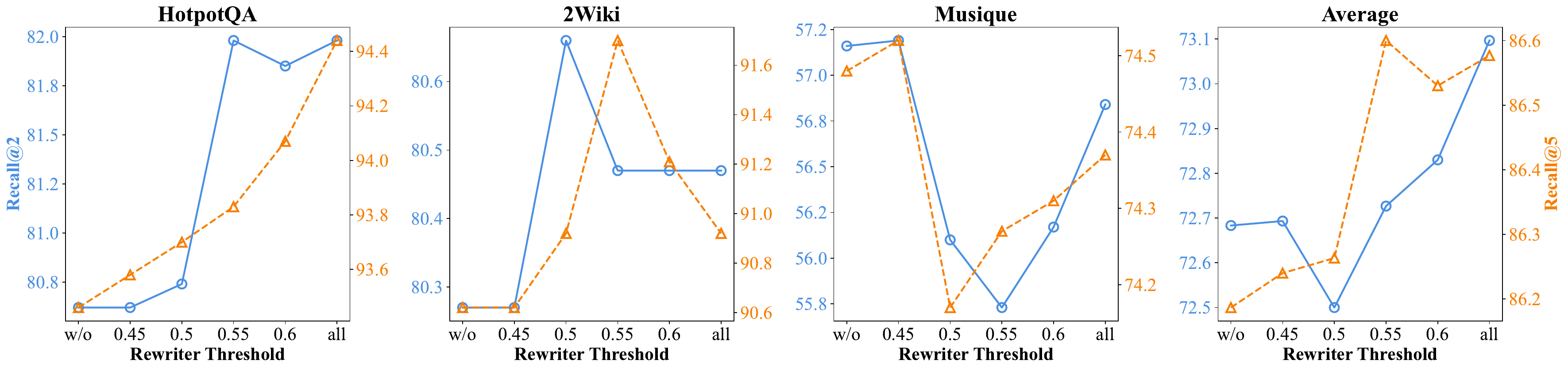}
    \caption{Effect of the rewriter threshold on retrieval performance across datasets (integrated with HippoRAG 2).}
    \label{fig:rewriter_thresholds_recall}
\end{figure*}
\subsection{\revise{Efficiency Analysis}}
To further validate the practical efficiency of \method{}, we conduct additional experiments focusing on latency and routing overhead.

\smallskip
\noindent \textbf{End-to-End Latency.} 
We report the end-to-end latency per query in~\Cref{fig:latency_decomposition}.
Always applying graph-based RAG leads to substantially higher latency than vanilla RAG, reflecting the cost of graph-based retrieval and reasoning.
By introducing routing, \method{}-Router consistently reduces latency compared to always using graph-based RAG when integrated with both GraphRAG and HippoRAG 2.
Compared to \method{}-Router, enabling rewriting introduces additional latency due to the use of an LLM rewriter.
However, the increase is moderate across datasets, indicating that the overhead of rewriting remains limited in practice.
At the same time, queries routed to vanilla RAG continue to benefit from substantially lower latency, which offsets the additional cost incurred on the rewritten subset.
These results suggest that the router alone already provides strong efficiency gains, while the full adapter achieves a better balance between effectiveness and efficiency with only a modest increase in latency.

\smallskip
\noindent \textbf{Routing Overhead.} 
\Cref{tab:routing time} and~\Cref{fig:router_bubbles} collectively illustrate the efficiency and effectiveness of different routers. Our \method{}-Router consistently reaches the performance upper bound, achieving the highest F1 scores across all three datasets. Notably, \method{}-Router outperforms the LLM-Router in accuracy while being approximately 4.6$\times$ faster, proving that a specialized adapter can surpass general-purpose LLMs in routing precision with significantly lower overhead.
Compared to the NER-Router, which exhibits uncertain routing preferences (e.g., over-routing in HotpotQA but under-routing in MuSiQue), \method{} demonstrates more stable identification of complex queries. While NER-Router is the fastest (7ms), the marginal latency increase of \method{} (16ms) yields substantial F1 gains, i.e., 3.9\% on average, positioning it as the most cost-effective solution for balancing system throughput and answer quality.

\subsection{Impact of Query Rewriting on Retrieval}
To investigate how query rewriting affects the retrieval quality of graph-based RAG, we evaluate the Recall@2 and Recall@5 across different rewriting thresholds, as shown in~\Cref{fig:rewriter_thresholds_recall}. While the results show a consistent upward trend in average recall, the overall magnitude of the improvement remains relatively marginal, i.e., a gain of less than 1.5\% in most settings. This suggests that the primary benefit of the rewriter is not merely to enhance the surface-level retrieval of evidence, but rather to facilitate the structural alignment between the query and the knowledge graph. By decomposing the query into explicit relational triplets without introducing external facts, the rewriter transforms implicit dependencies into a reasoning-oriented representation $q_r$. Consequently, the observed performance gains in downstream QA tasks likely stem from the ability of the rewriter to expose the latent multi-hop structure of the query, thereby providing a clearer logical roadmap for the graph-based reasoning module rather than significantly expanding the scope of retrieved entities.
\end{document}